\documentclass[letterpaper]{article} 
\usepackage{aaai2026}  
\usepackage{times}  
\usepackage{helvet}  
\usepackage{courier}  
\usepackage[hyphens]{url}  
\usepackage{graphicx} 
\usepackage{natbib}  
\usepackage{amsmath}
\usepackage{booktabs} 
\usepackage{multirow} 
\usepackage{xcolor}         
\usepackage{amssymb}        
\usepackage{colortbl}       
\usepackage{makecell}       
\usepackage{pifont}
\usepackage{svg}
\definecolor{darkgreen}{RGB}{50,100,0}
\definecolor{darkred}{RGB}{200, 0, 0}
\newcommand{\cmark}{\textcolor{darkgreen}{\ding{51}}} %
\newcommand{\xmark}{\textcolor{darkred}{\ding{55}}} %

\usepackage{caption} 
\usepackage{algorithm}
\usepackage{algorithmic}

\usepackage{graphicx}
\usepackage{subcaption} 
\usepackage{newfloat}
\usepackage{listings}
\DeclareCaptionStyle{ruled}{labelfont=normalfont,labelsep=colon,strut=off} 
\floatstyle{ruled}
\newfloat{listing}{tb}{lst}{}
\floatname{listing}{Listing}
\nocopyright

\title{Sharp Eyes and Memory for VideoLLMs: Information-Aware Visual Token Pruning for Efficient and Reliable VideoLLM Reasoning}
\author{
    Jialong Qin\textsuperscript{\rm 1,2,3},
    Xin Zou\textsuperscript{\rm 1,2},
    Di Lu\textsuperscript{\rm 1},
    Yibo Yan\textsuperscript{\rm 1,2},  
    Xuming hu\textsuperscript{\rm 1,2}\thanks{Corresponding Author}
}
\affiliations{
    \textsuperscript{\rm 1}The Hong Kong University of Science and Technology (Guangzhou)\\
    \textsuperscript{\rm 2}The Hong Kong University of Science and Technology\\
    
    \textsuperscript{\rm 3}Beijing Institute of Technology\\
    qinjialong5@gmail.com, xuminghu97@gmail.com\\

}

\usepackage{bibentry}

\begin{document}

\maketitle

\begin{abstract}
Current Video Large Language Models (VideoLLMs) suffer from quadratic computational complexity and key-value cache scaling, due to their reliance on processing excessive redundant visual tokens. To address this problem, we propose SharpV, a minimalist and efficient method for adaptive pruning of visual tokens and KV cache. Different from most uniform compression approaches, SharpV dynamically adjusts pruning ratios based on spatial-temporal information. Remarkably, this adaptive mechanism occasionally achieves performance gains over dense models, offering a novel paradigm for adaptive pruning. During the KV cache pruning stage, based on observations of visual information degradation, SharpV prunes degraded visual features via a self-calibration manner, guided by similarity to original visual features. In this way, SharpV achieves hierarchical cache pruning from the perspective of information bottleneck, offering a new insight into VideoLLMs' information flow. Experiments on multiple public benchmarks demonstrate the superiority of SharpV. Moreover, to the best of our knowledge, SharpV is notably the first two-stage pruning framework that operates without requiring access to exposed attention scores, ensuring full compatibility with hardware acceleration techniques like Flash Attention.
\end{abstract}

\section{Introduction}

\begin{table}[!ht]
  \centering
  \scriptsize 
  \setlength{\tabcolsep}{1.5pt} 
  \begin{tabular}{lccccc}
    \toprule[0.12em]
    \makecell{\textbf{Methods}\\\textit{Pre-LLM Stage}}   &  
    \makecell{\textbf{Granularity}\\\textbf{Level}} & 
    \makecell{\textbf{CLS}\\\textbf{Independent}} &
    \makecell{\textbf{Adaptive}\\\textbf{Ratio}} &
    \makecell{\textbf{Time}\\\textbf{Complexity}}\\
    \midrule
    DivPrune  {\tiny{(CVPR25)}}   
    &\textbf{Image} &\cmark & \xmark &$O(n^2 \cdot d + n^3)$\\
    PruMerage {\tiny{(ICCV25)}}    
    &\textbf{Image} &\xmark &\xmark &$O(n \cdot d)$\\
    PruneVid$^\ast$ {\tiny{(ACL25)}}  
    &\textbf{Segment} &\cmark &\xmark &$O(n^2\cdot d+n \cdot d)$\\
    DyCoke$^\ast$ {\tiny(CVPR25)}      
    &\textbf{Frame} &\cmark &\xmark &$O(n \cdot d)$\\
    VidCom\textsuperscript{2} {\tiny(2025.05)}  
    &\textbf{Frame} &\cmark&\xmark &$O(n \cdot d)$\\
    \rowcolor{blue!8} 
    \textbf{SharpV$^\ast$} {(ours)} 
     &\textbf{Frame} & \cmark & \cmark &$O(n \cdot d)$\\
     \midrule
     \makecell{\textbf{Methods}\\\textit{Intra-LLM Stage}}     &  
    \makecell{\textbf{Flash}\\\textbf{Attention}} &
    \makecell{\textbf{Adaptive}\\\textbf{Layer}} &
    \makecell{\textbf{Cache}\\\textbf{Reduction}}&
    \makecell{\textbf{Space}\\\textbf{Complexity}}\\
    \midrule
    FastV {\tiny{(ECCV24)}}      
    &\xmark &\xmark &\cmark &$O(n^2 \cdot d)$\\
    VTW  {\tiny{(CVPR25)}}   
    &\xmark & \cmark & \cmark &$O(n^2 \cdot d)$ \\
    PruneVid$^\dagger$  {\tiny{(ACL25)}}  
    &\xmark  &\xmark &\cmark &$O(n^2 \cdot d)$\\
    DyCoke$^\dagger$ {\tiny(CVPR25)}      
    &\xmark &\xmark  &\cmark &$O(n^2 \cdot d)$ \\
    FrameFusion {\tiny(ICCV25)}      
    &\xmark &\xmark  &\cmark &$O(n^2 \cdot d)$ \\
    \rowcolor{pink!35} 
    \textbf{SharpV$^\dagger$} {(ours)} 
    &\cmark &\cmark & \cmark &$O(n \cdot d)$\\
    \bottomrule[0.12em]
  \end{tabular}
    \caption{\small Why SharpV Outperforms: A Systematic Comparison.
  Where $^\ast$ denotes \textit{Pre-LLM} pruning, $^\dagger$ indicates \textit{Intra-LLM} pruning.}\vspace{-0.85em}
  \label{tab:feature_comparison}
\end{table}

Video Large Language Models (VideoLLMs) have demonstrated remarkable capabilities in video understanding and reasoning tasks \cite{videochat,llavanextvideo,llavavideo,videollava}. However, the extensive spatiotemporal information in long videos consumes a significant portion of LLM input tokens \cite{llavaov,pllava}, leading to excessive computational overhead and memory usage \cite{slowfast}. As shown in Table \ref{tab:feature_comparison}, existing pre-LLM stage approaches have explored visual token pruning at various granularity levels to reduce token count and memory consumption. However, these methods \cite{fastv,vtw,sttm,dycoke,framefusion} uniformly apply fixed pruning ratios, which typically represent local optima tuned for specific datasets, lacking adaptive pruning based on video information content. This limitation affects their generalization and robustness. Moreover, since the primary objective of visual pruning is computational efficiency, the pruning process itself should maintain minimal complexity to avoid additional overhead. Existing approaches that rely on computationally intensive clustering algorithms \cite{prunevid} or mathematically sophisticated planning techniques \cite{llavascissor,divprune} may inadvertently introduce non-trivial overhead to the pruning process shown in Table~\ref{tab:feature_comparison}, potentially limiting their practical applicability in real-world deployment scenarios where computational efficiency is paramount. To address these challenges, in pre-LLM stage, we propose a novel adaptive pruning method (Visual SharpV) that computes spatiotemporal importance for each token and determines frame-specific pruning ratios based on information volume, achieving fine-grained information preservation while maintaining low complexity.

In the intra-LLM stage, as demonstrated in Table~\ref{tab:feature_comparison}, most existing works reduce cache by analyzing exposed attention scores. For instance, \cite{fastv,vtw} retain visual tokens with high attention scores in specific layers based on inefficient visual attention phenomena, while \cite{prunevid} preserves tokens with high cross-attention scores. Although attention-based analysis provides reasonable motivation and valuable insights into attention mechanism dynamics, the emergence of efficient attention computation methods like flash attention \cite{flashattention,flashattention2}, which reduces the quadratic complexity of native attention to linear complexity while not exposing attention scores, renders such approaches incompatible. Therefore, we propose an information-theoretic perspective for intra-LLM stage pruning by measuring similarity between decoder layer inputs and original visual features to dynamically retain visual tokens for subsequent generation steps. Our method reduces complexity from $O(n^2 \cdot d)$ to $O(n \cdot d)$, fulfilling the lightweight design principle and the core objective of pruning.

We integrate SharpV into two widely adopted VideoLLMs with varying capabilities: PLLaVA \cite{pllava}, LLaVA-OneVision \cite{llavaov}. Extensive evaluations are conducted on four public datasets covering various video types (long, medium, short), including MVBench \cite{mvbench}, VideoMME \cite{videomme}, NExTQA\cite{nextqa}, ActNet-QA\cite{activitynet}, with comparisons against recent SOTA methods and baselines. Our experiments demonstrate the effectiveness of information-aware adaptive pruning ratios: at similar or lower retention ratios, SharpV achieves higher accuracy than SOTA methods while delivering more significant speedups due to its extremely low computational overhead.

\begin{figure*}[t]
\centering
\includegraphics[width=\linewidth]{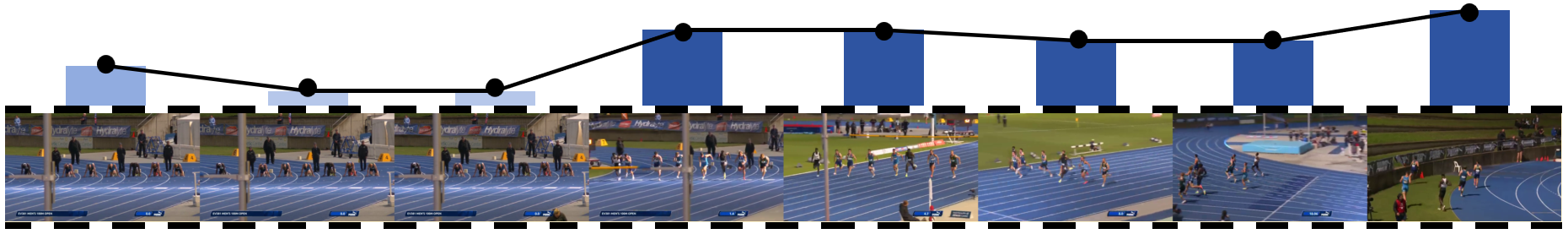} 
\caption{VideoLLM Response Demonstration Using Training-Free SharpV Pruning, Blue bars denote the per-frame token retention ratios dynamically computed by Visual SharpV, reflecting information-aware pruning.}
\label{fig:threshold}
\end{figure*}
Our main contributions are threefold:
\begin{itemize}
    \item For the pre-LLM stage, we propose information-aware visual token pruning that maximize useful information retention at both frame and token levels, achieving video-specific adaptive pruning while maintaining lightweight $O(n \cdot d)$ complexity to accelerate VideoLLM inference.
    
    \item Departing from conventional attention-based analysis, we introduce an information-theoretic perspective for intra-LLM stage pruning. From the perspective of mutual information, our method is the first framework totally compatible with flash attention, reducing memory overhead, and offering new insights into LLM information flow.
    
    \item Comprehensive experiments on public datasets demonstrate SharpV's consistent performance and efficiency advantages. Our method consistently achieves a competitive trade-off between simplicity, efficiency, and accuracy compared to other SOTA approaches.
\end{itemize}

\section{Related Work}

\subsection{The Intrinsic Redundancy in Videos}
Videos inherently exhibit spatio-temporal locality, which has been exploited in deep learning approaches. Early works like 3D CNNs \cite{3dcnn1,3dcnn2} and VideoTransformers \cite{tong2022videomae,spatiotemporallearners} achieved promising results in feature learning by leveraging this redundancy. Similarly, in multimodal LLM inference scenarios, visual token compression methods have enabled efficient model reasoning. Some works employ various strategies to identify spatially important tokens, typically by filtering tokens based on features from visual encoders \cite{dynamicvit,fastvit,tome,spvit}. For instance, PruneMerge \cite{prumerge} selects crucial tokens based on inherent attention characteristics \cite{clipattention} between [CLS] token and other visual token. Other works rely on similarity-based methods to identify important tokens. DivPrune \cite{divprune} reduces spatial redundancy by solving for the maximally diverse set that maintains minimum pairwise distances. Cross-modal interaction methods like SimIgnore \cite{simignore} filter instruction-relevant visual tokens by comparing their similarity with prompt tokens. However, these approaches only consider spatial redundancy while neglecting temporal redundancy in videos.

For spatio-temporal video compression, LLaVA-Scissor \cite{llavascissor} preserves semantic information through two-stage connectivity analysis, but suffers from high computational costs due to graph construction and Union-Find algorithms. PruneVid \cite{prunevid} focuses on compressing static backgrounds but requires video segmentation via clustering, which is computationally intensive and struggles with dynamic parameter settings. VidCom\textsuperscript{2} \cite{vicom2} quantifies frame uniqueness through global token representations, which produces excessively coarse-grained features and proves inadequate for real-time inference scenarios. Dynamic methods like DyCoKe \cite{dycoke} and Frame Fusion \cite{framefusion} efficiently compress inter-frame tokens through similarity measures, yet face challenges in setting adaptive thresholds for diverse video types. 

In contrast, SharpV introduces frame-specific dynamic pruning without manual intervention, maintaining low latency while avoiding additional computational overhead.

\subsection{Intra-LLM Visual Token Pruning}
During prefilling and decoding phases of multimodal large models, most approaches reduce computational costs and KV cache by analyzing attention distributions. FastV \cite{fastv} and VTW \cite{vtw} discovered that visual tokens receive significantly diminished attention in deeper layers, enabling pruning or removal in late stages. FitPrune \cite{fitprune} formulates pruning as a statistical distribution fitting problem, minimizing attention distribution discrepancies pre- and post-pruning. PrunerVid \cite{prunevid} performs uniform layer-wise pruning based on cross-attention scores between prompts and visual tokens, effectively reducing KV cache. DyCoKe \cite{dycoke} designs a special DP Cache to retain potentially useful tokens according to attention scores during subsequent decoding stages, optimizing matrix operations. Frame Fusion \cite{framefusion} accumulates cross-layer attention scores to measure token importance.

While attention-based pruning is theoretically reasonable, modern hardware acceleration strategies like flash attention \cite{flashattention,flashattention2} have reduced quadratic complexity to linear complexity at the cost of hiding attention scores, rendering many existing pruning methods obsolete. Addressing this challenge, SharpV pioneers a novel pruning paradigm that operates independently of exposed attention scores, providing fresh insights for visual token compression.

\section{Methodology}

\begin{figure*}[t]
\centering
\includegraphics[width=\linewidth]{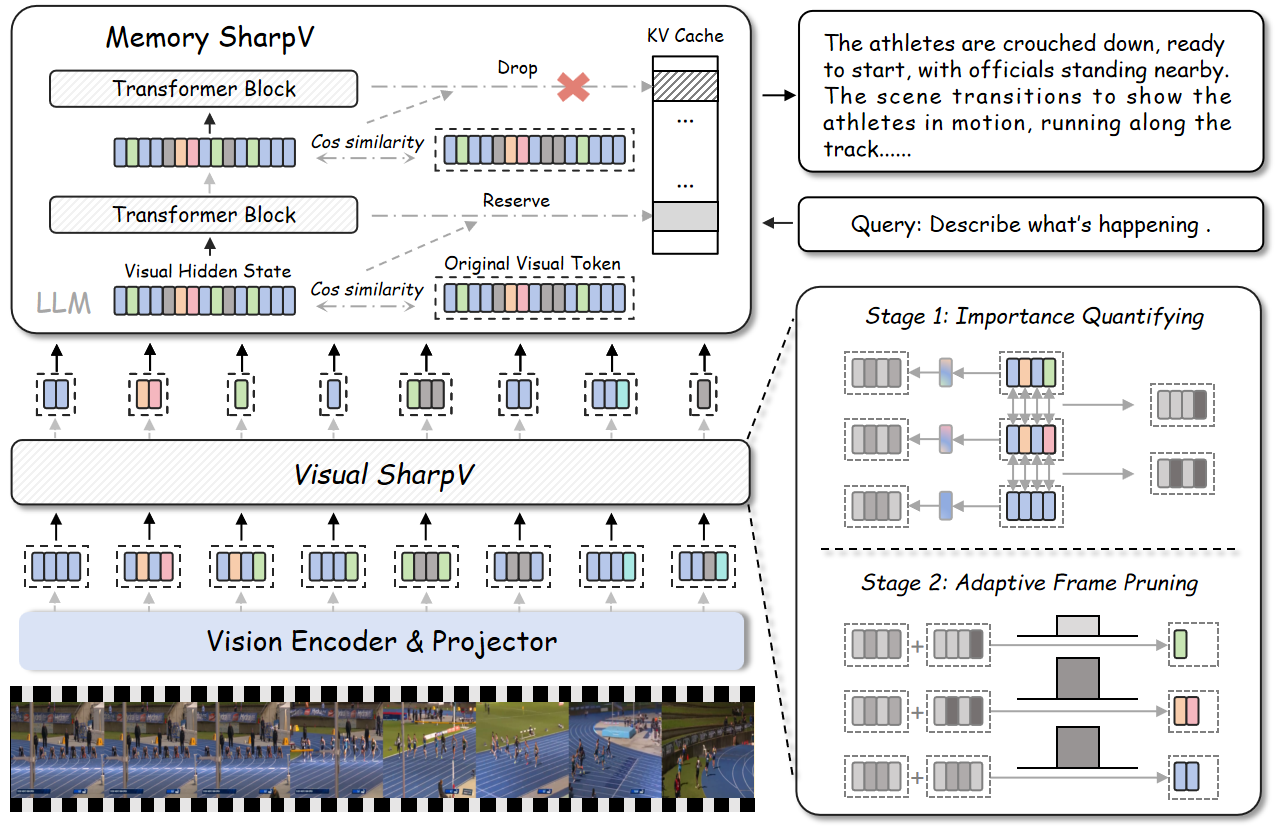} 
\caption{\textbf{The Detailed Overview of SharpV.} SharpV is a two-stage training-free plug-and-play framework for video LLM pruning. In the pre-LLM stage, Visual SharpV selects important visual tokens based on spatio-temporal scores, with an adaptive pruning ratio determined by L2 norm and a dissimilarity module. In the intra-LLM stage, Memory SharpV dynamically discards key-value cache by evaluating layer-wise visual information degradation.}

\label{fig:main}
\end{figure*}

\subsection{Information-Aware Visual Space Pruning}
Our visual space pruning framework (Visual SharpV) operates through two stages: For a video sampling $n$ frames, first, we evaluate the intra-frame token importance to obtain the spatial dimension scores of $n$ frames and determine the pruning ratio for the first frame. Then, we analyze motion patterns within the temporal sequence to assess inter-frame information variation, computing temporal dimension scores and pruning ratios for the remaining $n-1$ frames. Finally, by aggregating each token's spatial and temporal scores, we retain high-scoring tokens according to the dynamically calculated per-frame ratios, resulting in an information-aware visual space representation.

\subsubsection{Dissimilarity Computation Module} 

As is well-known, cosine similarity effectively measures directional alignment between vectors. In this work, we aim to quantify token dissimilarity as a positive measure. While some approaches simply rescale cosine similarity using (1 - cosine similarity), this becomes problematic in high-dimensional spaces \cite{cosdim1,cosdim2}, where random vectors have expected cosine similarity approaching zero \cite{cosine}.

We therefore propose SharpV Dissimilarity, which measures directional differences using Euclidean distance between unit vectors. This metric amplifies subtle directional discrepancies between nearly-aligned vectors:
\begin{equation}
\text{Dissim}(\mathbf{v}_1, \mathbf{v}_2) = \|\hat{\mathbf{v}_1} - \hat{\mathbf{v}_2}\|_2 = \sqrt{2 - 2\cos (\theta)},
\end{equation}
where $\hat{\mathbf{v}} = \frac{\mathbf{v}}{\|\mathbf{v}\|_2}$ denotes L2-normalized vectors and $\theta$ is their angular separation. This paper employs Dissim to measure all visual token dissimilarities. We offer this module to prepare for a self-calibrated adaptive pruning ratio. Notably, replacing Dissim with (1 - cosine similarity) and a manually set fixed similarity threshold $K$ is also compatiable with SharpV framework.

\subsubsection{Spatio-Temporal Token Importance}
While some approaches compute pairwise token similarity to preserve critical semantic information \cite{divprune,llavascissor}, this inevitably leads to quadratic computational complexity. Unlike existing methods \cite{vicom2} that coarsely estimate frame-level uniqueness, we propose a novel token-level spatio-temporal importance assessment mechanism. Assisted by our designed Dissimilarity Computation Module, each token's importance is comprehensively evaluated along dual dimensions.

For VideoLLMs, videos are sampled into $n$ frames, each frame is mapped into $f$ visual tokens. VideoLLMs take $n \times f$ visual tokens as complete visual input. For the tokens within each individual frame, only spatial information exists. In the spatial domain, the dissimilarity between each visual token and the overall spatial representation constitutes its Spatial Importance. Therefore, in a straightforward approach, we quantify the uniqueness of each token by measuring its dissimilarity with the frame's average representation. Given the set of tokens $F_t \in R^{f×d}$ for frame t, where the global average representation is denoted as $\overline{F_t} \in \mathbb{R}^d$, the Spatial Importance $\mathcal{S}$ for each token is defined as:
\begin{equation}
\mathcal{S} = \textbf{Dissim}(F_t, \overline{F_t}) \in \mathbb{R}^f,
\end{equation}
where $\overline{F_t}$ is unsqueezed along the $0th$ dimension before being input into the dissimilarity module.

The temporal importance we defined for visual tokens is calculated through the following steps: Firstly, we reshape the token matrix $V \in \mathbb{R}^{N_v \times d}$ into a three-dimensional matrix $\mathbb{R}^{n \times f \times d}$ where $d$ is the token dimension and $n \times f = N_v$. Subsequently, temporal importance $\mathcal{T}$ is calculated by dissimilarity between corresponding tokens in adjacent frames:
\begin{equation}
\mathcal{T} = \textbf{Dissim}(\mathbf{F}_{t},\mathbf{F}_{t-1}) \in \mathbb{R}^{f}
\quad t \in \mathbb{Z} \cap [1,n],
\end{equation}
where $\mathbf{F}_{t},\mathbf{F}_{t-1} \in \mathbb{R}^{f\cdot d}$ are the token set of frame $t$ and $t-1$.

The final importance score $\mathcal{I}$ of each token is computed by jointly considering its spatial and temporal importance:
\begin{equation}
    \mathcal{I} = \mathcal{T} + w\cdot\mathcal{S},
\end{equation}
Since the distributions of variables $\mathcal{T}$ and $\mathcal{S}$ are different, we introduce a hyperparameter $w$ to control the incorporation ratio of $\mathcal{S}$. By jointly modeling and aggregating temporal and spatial token importance, Visual SharpV provides more comprehensive token evaluation than single-dimension approaches.

\subsubsection{Information-Aware Threshold Pruning}
In this section, we offer a lightweight adaptive threshold strategy that dynamically adjusts token retention rates frame-by-frame. This approach computes inter-frame variation without additional parameters and preserves only regions exhibiting meaningful changes, achieving fine-grained pruning at frame-level granularity. The adaptive mechanism automatically increases token retention for high-motion sequences while aggressively pruning static frames.

Since the Euclidean norm of a frame $t$'s temporal importance $\|\mathcal{T}_t\|_2 \in [0, 2\sqrt{f}]$, the Euclidean distance of a frame's temporal importance quantifies its visual variation relative to the preceding frame. The upper bound $2\sqrt{f}$ represents maximal inter-frame variation, while zero temporal importance indicates identical frames, corresponding to static visual content. Hence, we calculated $\text{threshold}_{t}$ for frame $t\in \mathbb{Z} \cap [2,n]$ as follow:
\begin{equation}
\text{threshold}_{t} = \frac{\|\mathcal{T}_t\|_2}{2\sqrt{f}}\in [0,1].
\end{equation}
For the first frame, since temporal importance is not applicable as no preceding frames exist, we maintain consistency by computing its threshold in a manner analogous to the subsequent $n-1$ frames. Specifically:
\begin{equation}
\text{threshold}_{1} = \frac{\|\mathcal{S}_1\|_2}{2\sqrt{f}}\in [0,1].
\end{equation}
This is equivalent to virtually supplementing a complete reference frame where all tokens are represented by $\overline{F_1}$ as the preceding frame for temporal threshold computation of the first frame.

As illustrated in Fig.~\ref{fig:threshold}, our Information-Aware Threshold Strategy dynamically adjusts pruning thresholds by monitoring token norm variations, enabling frame-wise adaptive pruning through top-K selection of token importance scores $\mathcal{I}$. Detailed overview is shown in Fig.~\ref{fig:main}.

\subsection{Information-Aware Memory Space Pruning}
\begin{figure}[t]
\centering
\begin{subfigure}[b]{\linewidth}
    \includegraphics[width=1\linewidth]{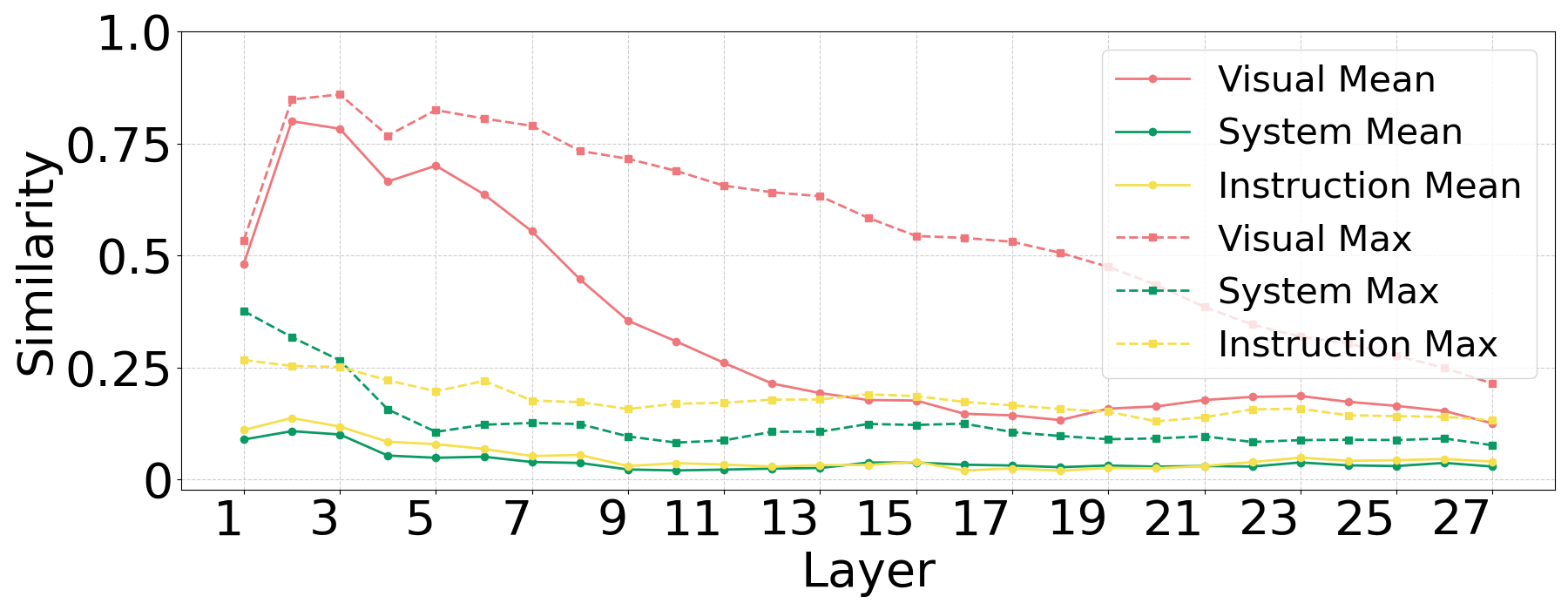}
    \caption{Llava-OneVision-7B}
    \label{subfig:simiov}
\end{subfigure}
\hfill
\begin{subfigure}[b]{\linewidth}
    \includegraphics[width=\linewidth]{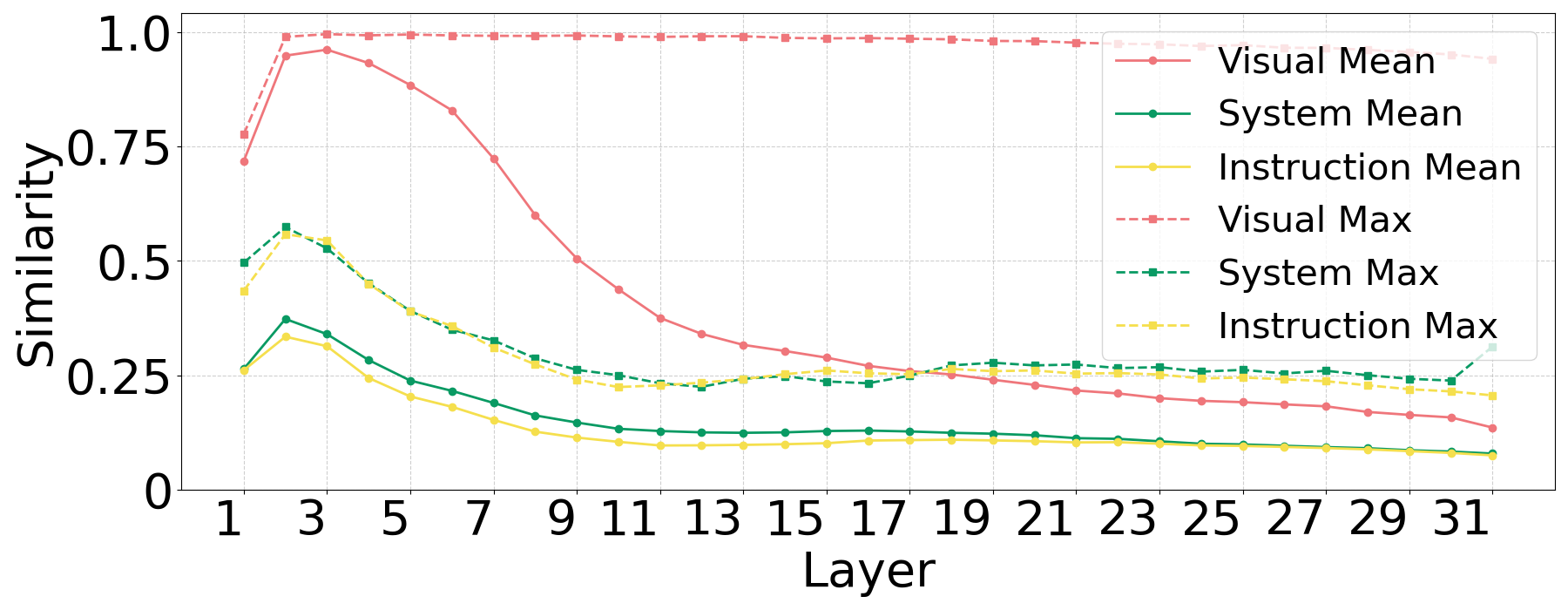}
    \caption{PLLaVA-7B}
    \label{subfig:simipllava}
\end{subfigure}
\caption{Similarity across different layers}
\label{fig:simi}
\end{figure}

\subsubsection{Visual Information Degradation Hypothesis}

As shown in Fig.~\ref{fig:simi}, we comprehensively compare the cosine similarity between hidden states $G_l = (S_l,V_l,I_l)\in \mathbb{R}^{(N_s + N_v + N_p) \times d}$ at different layers $l$ and their original positions in $G = (S,V,I)\in \mathbb{R}^{(N_s + N_v + N_p) \times d}$ for both LLaVA-OneVision-7B and PLLaVA, visualizing their average values. A striking observation is that in both models, the visual tokens $V_l$ exhibit high similarity on average in shallow layers and demonstrate a clear decreasing trend to their original counterparts $V$ as layer depth increases. In contrast, system and instruction tokens rapidly converge to near-zero similarity (0-0.2 range). Mathematically, the similarity of random high-dimensional vectors approaches zero, suggesting that system and instruction tokens undergo more complex unknown transformations through Attention and MLP operations. Meanwhile, visual tokens show measurable degradation of original visual information via cosine similarity, which we term \textit{Visual Information Degradation}. Although the exact relationships between transformed token embeddings remain unclear \cite{simignore}, the similarity patterns reveal that visual information undergoes rapid decay in shallow layers resembling human memory curves before stabilizing at low similarity levels (<0.2) after layer 14. This suggests LLMs primarily process visual information in early layers, with diminishing importance in deeper layers. 

\subsubsection{Insights}
Research on visual attention reveals a consistent pattern: Inefficient Visual Attention phenomenon \cite{fastv,vtw} validate that visual tokens receive drastically diminished attention scores in deeper layers of multimodal LLMs. We further quantify this via cosine similarity, a robust metric for high-dimensional feature decay \cite{cosine2,cosine3,cosine4,cosine5}.

From the perspective of information theory, this degradation pattern aligns with the concept of progressive information compression in hierarchical systems. The rapid similarity decay in early layers suggests intensive feature extraction, while the stable low similarity in deeper layers indicates visual information has been sufficiently encoded into subsequent representations. This mirrors the information bottleneck principle \cite{bottleneck,videobottleneck}, where networks discard irrelevant details while preserving task-relevant features through successive transformations \cite{deepbottleneck}. The distinct behavior of visual versus textual tokens further implies modality-specific information \cite{crossmodal1,crossmodal2} processing pathways in multimodal LLMs, analogous to attentional load theory in human perception \cite{visualattention1}.

\subsubsection{Degradation-Aware Pruning}

Building upon this observation, we propose a simple yet effective strategy of discarding degraded visual information. Specifically, when the cosine similarity between a layer's visual tokens and the original visual information falls below a threshold $M$, we discard the corresponding layer's KV Cache. Formally,
\begin{equation}
    \text{Discard}(l) = 
    \begin{cases} 
    \text{True}, & \text{if } \cos (\mathbf{V}_l, \mathbf{V}) < M \\
    \text{False}, & \text{otherwise}
    \end{cases},
\end{equation}
where $\mathbf{V}_l$ represents the visual tokens at layer $l$, $\mathbf{V}$ denotes the original visual information and $M$ is defined as the degradation threshold. When $\text{Discard}(l) = \text{True}$, the KV cache of layer $l$ is discarded to prevent degraded features from propagating further. This ensures that only high-quality visual information is retained for subsequent decoding.

\begin{table}[t]
\centering
\small
\begin{tabular}{lcccc}
\toprule
Model & Dimension & Layer & \#Tokens & \#Frames \\
\midrule
PLLaVA-7B & 4096  & 32 & 144 & 16 \\
LLaVAOV-0.5B & 896  & 24 & 196 & 32 \\
LLaVAOV-7B & 3584 & 28 & 196 & 32 \\
\bottomrule
\end{tabular}
\caption{Model Specifications}\vspace{-0.5em}
\label{Model Specifications}
\end{table}

\definecolor{mvbench}{RGB}{255, 255, 255}
\definecolor{videomme}{RGB}{255, 255, 255}  
\definecolor{nextqa}{RGB}{255, 255, 255} 
\definecolor{avg}{RGB}{255, 255, 255}

\definecolor{graybg}{RGB}{240, 240, 240}     
\definecolor{sharpv1}{RGB}{245, 241, 253}     
\definecolor{sharpv}{RGB}{235, 232, 248} 

\begin{table*}[!t] 
\centering
\resizebox{\textwidth}{!}{
\begin{tabular}{l|c|cc*{3}{>{\centering\arraybackslash}p{1.2cm}}|*{1}{>{\centering\arraybackslash}p{1.25cm}}|*{1}{>{\centering\arraybackslash}p{1.4cm}}*{1}{>{\centering\arraybackslash}p{2.1cm}}*{3}{>{\centering\arraybackslash}p{1.4cm}}}
\toprule
\multirow{2}{*}{\textbf{Method}} & 
\multirow{2}{*}{\makecell{\textbf{Token}\\\textbf{Budget}}} & 
\multicolumn{1}{c}{\cellcolor{mvbench}\textbf{MVBench}} & 
\multicolumn{1}{c}{\cellcolor{videomme}\textbf{VideoMME}} & 
\multicolumn{1}{c}{\cellcolor{mvbench}\textbf{NextQA}} & 
\multicolumn{2}{c|}{\cellcolor{nextqa}\textbf{ActNet-QA}} & 
\multicolumn{1}{c|}{\cellcolor{avg}\textbf{Avg.}} & 
\multicolumn{5}{c}{\textbf{Efficiency Analysis}}\\
\cmidrule(lr){3-3} \cmidrule(lr){4-4} \cmidrule(lr){5-5} \cmidrule(l){6-7}\cmidrule(l){8-13}
& & Acc. $\uparrow$ & wo $\uparrow$ & mc $\uparrow$ & Acc. $\uparrow$ & Sco. $\uparrow$ & Acc. $\uparrow$ & Memory & TTFT $\downarrow$ & VR $\downarrow$ & MR $\downarrow$ & TFlops \\

\midrule
\rowcolor{graybg}
LLaVA-OneVision 0.5B &100\% & 46.6 & 45.9 & 57.5 & 47.9 & 2.66 & 49.5 & 18G & 0.61s (1.00$\times$) & 100\% & 100\% & 3.4 \\
\midrule
+ FastV &30\% & 44.7 & 40.9 & 55.8 & 46.3 & 2.53 & 46.9 & 14G & 0.49s (1.24$\times$) & 100\% & 30\% & 1.4 \\
+ PruMerge &55\% & 42.5 & 38.8 & 54.5 & 41.7 & 2.35 & 44.4 & 13G & 1.38s (0.44$\times$) & 55\% & 100\% & 1.5 \\
+ DyCoke  &19\% & 46.3 & 45.1 & 57.2 & 47.8 & 2.65 & 49.1 & 12G & 0.46s (1.33$\times$) & 50\% & 38\% & 1.8 \\
+ DyCoke  &15\% & 46.5 & 45.2 & 57.5 & 47.7 & 2.65 & 49.2 & \underline{10G} & 0.41s (1.49$\times$) & 30\% & 48\% & \underline{1.2} \\
\rowcolor{sharpv1}
\textbf{+ SharpV (Manual)} &19\% & \textbf{46.7} & \underline{46.7}  & \textbf{57.8} & \textbf{47.9} & \textbf{2.66} & \textbf{50.4} & \underline{10G} & \underline{0.39s} (1.56$\times$) & 49\%  & 39\%  & 1.7\\

\rowcolor{sharpv}
\textbf{+ SharpV (Adaptive)} & 12\% & \underline{46.6} & \textbf{46.9}  & \underline{57.7} & \underline{47.8} & \underline{2.65} & \underline{49.8} & \textbf{9G} & \textbf{0.37s} (1.65$\times$) & 32\%  & 39\% & \textbf{1.1}  \\
\midrule
\rowcolor{graybg}
LLaVA-OneVision 7B &$100\%$ & 57.7 & 58.7 & 79.1 & 51.9 & 2.86 & 61.9 & 32G & 1.05s (1.00$\times$) & 100\% & 100\% & 41.4 \\
\midrule
+ FastV &$30\%$ & 56.3 & 56.4 & 76.5 & 50.7 & 2.80 & 60.0 & 30G & 0.81s (1.30$\times$) & 100\% & 30\% & 17.9 \\

+ PruMerge &55\% & 52.2 & 52.9 & 75.0 & 50.5 & 2.78 & 57.7 & 28G & 1.73s (0.61$\times$) & 55\% & 100\% & 21.1 \\

+ DyCoke  &19\% & 57.7 &  59.3 & 78.4 & \underline{52.1} & \underline{2.88} & 61.9 & 28G & 0.77s (1.36$\times$) & 50\% & 37.5\% & 24.1 \\
+ DyCoke  &15\% & 57.2 &  58.3& 78.1 & 51.8 & 2.85 & 61.4 & 24G & 0.73s (1.44$\times$) & 30\% & 47.5\% & \underline{17.9} \\

\rowcolor{sharpv1}
\textbf{+ SharpV (Manual)} &19\% & \underline{57.8} & \underline{59.5}  & \textbf{79.0} & \textbf{52.0} & \textbf{2.87} & \underline{62.1} & \underline{23G} & \underline{0.67s} (1.57$\times$) & 49\%  & 39\%  & 23.9 \\
\rowcolor{sharpv}
\textbf{+ SharpV (Adaptive)} &12\% & \textbf{58.2} & \textbf{60.0}  & \underline{78.8} & 51.9 & 2.86 & \textbf{62.2} & \textbf{22G} &  \textbf{0.64s} (1.64$\times$) & 32\%  & 39\%  & \textbf{16.7}  \\

\bottomrule
\end{tabular}
}
\caption{Comparison of training-free visual token pruning methods using different VideoLLMs according to adaptive threshold pre-filling token budgets. Performance is measured by Accuracy (Acc), Computational Flops, and Time To First Token (TTFT).}
\label{tab:main}
\end{table*}

\section{Experiment}
\subsection{Evaluation Setting}
\subsubsection{Benchmarks}

We employ lmms-eval \cite{lmms} as our primary evaluation framework. For PLLaVA \cite{pllava} specifically, we use its official implementation to ensure fair comparison. Our comprehensive evaluation covers diverse video benchmarks, including MVBench \cite{mvbench} and VideoMME \cite{videomme} for multi-dimensional video understanding and NExT-QA \cite{nextqa} and ActNet-QA \cite{activitynet} for video question answering.

\subsubsection{Baselines}
We implement FastV \cite{fastv}, PruMerge \cite{prumerge} and DyCoKe \cite{dycoke} on two widely used LLMs for video understanding: PLLaVA\cite{pllava} and LLaVA-OneVision \cite{llavaov} with details shown in Tab.\ref{Model Specifications}. To demonstrate scalability, we evaluate 0.5B and 7B versions of LLaVA-OneVision. 

\subsubsection{Metrics} 
The term Token Budget combines the visual pruning ratio (VR) from the pre-LLM stage and the memory pruning ratio (MR) from the intra-LLM stage, defined as $VR \times MR$ following \cite{dycoke,framefusion}. VR measures the ratio of retained visual tokens fed into LLMs \cite{vicom2,divprune}, while MR estimates KV-Cache utilization during decoding \cite{framefusion}. For efficiency comparison, we evaluate FLOPs, GPU memory and acceleration metrics: Time To First Token (TTFT) and Time Per Output Token (TPOT).

\subsubsection{Implementation Details}
All experiments are conducted on NVIDIA A800 GPUs with 80GB memory. Since SharpV provides adaptive pruning strategies, we compare information-aware adaptive threshold $(M=0.2,w=1)$ with a semi-adaptive strategy using $(K=1.6,M=0.2,w=1)$. Token budgets are averaged across all datasets. Positional embedding undergoes re-encoding during intra-LLM pruning in Memory SharpV. For FastV, we set the selected layer to 3 and the pruning ratio to 0.3, following the official implementation. For DyCoKe, we adopt the hyperparameters (K, L, P) as (0.5, 3, 0.7) and (0.7, 3, 0.7), consistent with the original paper.
\begin{figure}
    \centering
    \includegraphics[width=\linewidth]{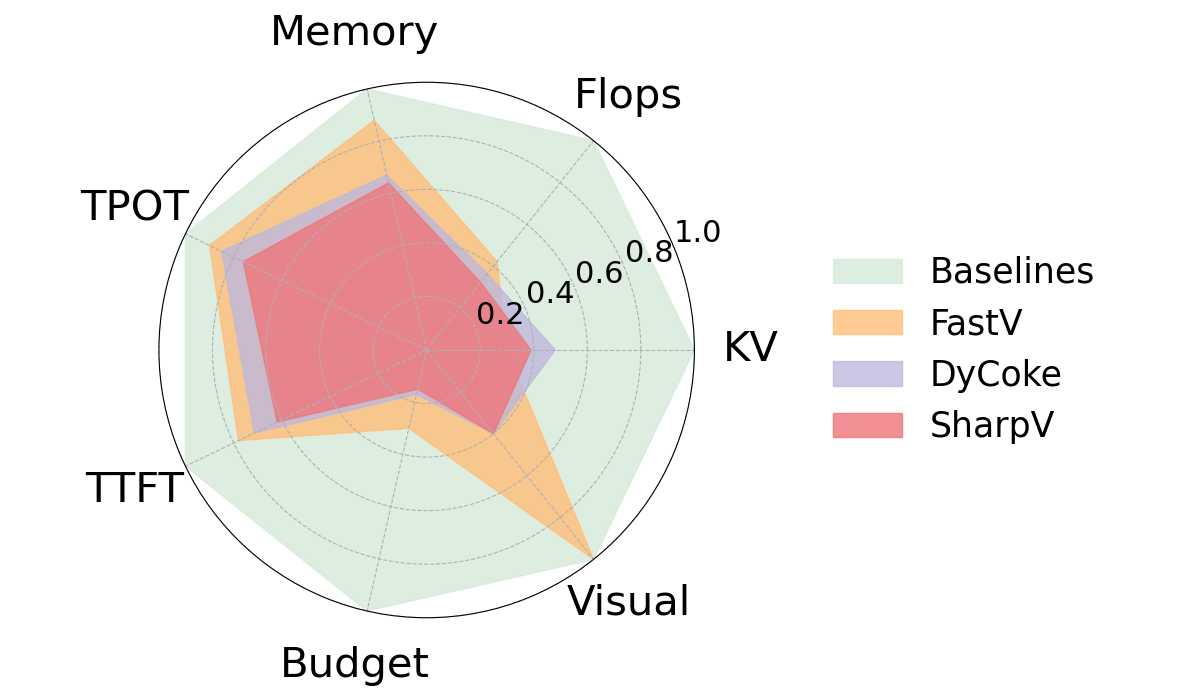}
    \caption{Average Efficiency}
    \label{fig:radar}
\end{figure}

\begin{table*}[t]
\centering
\resizebox{\textwidth}{!}{
\begin{tabular}{l|cccccccccccccccccccc|ccc|c}
\toprule
\multirow{2}{*}{\textbf{Method}} & 
\multicolumn{20}{c|}{\cellcolor{mvbench}\textbf{MVBench}} & 
\multicolumn{3}{c|}{\cellcolor{videomme}\textbf{VideoMME}} & 
\cellcolor{avg}\textbf{Avg.} \\
\cmidrule(lr){2-21} \cmidrule(lr){22-24} \cmidrule(lr){25-25}

& AA & AC & AL & AP & AS & CO & CI & EN & ER & FA & FP & MA & MC & MD & OE & OI & OS & ST & SC & UA & Short & Medium & Long & ACC. \\
\midrule
\rowcolor{graybg}
PLLaVA 7B & 55.5 & 39.5 & 26.0 & 49.0 & 58.0 & 53.5 & 31.0 & 30.5 & 48.0 & 41.0 & 42.0 & 52.0 & 45.0 & 23.5 & 56.0 & 61.0 & 36.0 & 82.0 & 42.0 & 61.0 & 51.4 & 43.4 & 35.4 & 46.6/43.4 \\
\midrule
\rowcolor{sharpv}
Visual SharpV & \textbf{62.0} & \textbf{41.0} & \textbf{27.5} & \textbf{50.5} & \textbf{56.5} & \textbf{53.5} & \underline{32.0} & \textbf{32.0} & \textbf{47.5} & \textbf{42.5} & \textbf{49.5} & \textbf{56.0} & \textbf{39.5} & \textbf{20.5} & \textbf{57.0} & \textbf{64.0} & \textbf{37.0} & \textbf{82.5} & \textbf{43.5} & \textbf{65.0} & \textbf{52.1} & \textbf{42.2} & \textbf{36.5} & \textbf{48.0/43.6} \\
V-Random$^f$ & \textbf{62.0} & \underline{38.5} & \underline{26.0} & \underline{50.0} & \underline{56.0} & \underline{47.0} & \textbf{37.0} & \underline{31.5} & \textbf{47.5} & \underline{41.5} & 40.5 & \underline{52.0} & \underline{39.0} & \underline{20.0} & 53.0 & \underline{60.5} & 34.0 & \underline{82.0} & \textbf{43.5} & \underline{63.5} & \underline{50.3} & \underline{40.4} & \underline{35.2} & \underline{46.3/42.0} \\
V-Random$^s$ & 61.0 & 38.0 & 24.0 & 49.5 & 55.0 & 45.0 & 29.0 & \underline{31.5} & 47.0 & 39.0 & \underline{42.0} & 51.0 & 37.5 & 18.0 & \underline{54.0} & 59.5 & \underline{34.5} & 81.5 & 42.0 & 60.5 & 49.4 & 40.2 & 34.3 & 45.0/41.3 \\

\midrule

\rowcolor{graybg}
LLaVA-OV 7B & 78.0 & 47.5 & 59.0 & 58.0 & 71.0 & 72.0 & 50.0 & 33.0 & 55.0 & 47.0 & 54.0 & 70.0 & 48.5 & 21.0 & 47.5 & 82.5 & 35.5 & 92.5 & 52.0 & 80.0 & 71.0 & 57.1 & 48.1 & 57.7/58.7 \\
\midrule
\rowcolor{sharpv}

Visual SharpV & \textbf{81.0} & \textbf{46.5} & \textbf{58.0} & \textbf{59.0} & \underline{75.0} & \underline{68.0} & \textbf{48.0} & \textbf{34.0} & \textbf{53.5} & \textbf{45.0} & \textbf{58.0} & \underline{62.5} & 46.0 & \textbf{28.0} & \underline{57.0} & \textbf{83.0} & 36.0 & \textbf{93.5} & \textbf{52.0} & \textbf{79.0} & \underline{70.8} & \textbf{57.4} & \textbf{50.2} & \textbf{58.2/59.5} \\
V-Random$^\dagger$ & \underline{79.5} & \underline{46.0} & \underline{47.5} & \underline{57.5} & \textbf{76.0} & \textbf{69.0} & \underline{39.5} & \underline{33.0} & \textbf{52.0} & \underline{44.5} & \underline{54.5} & \textbf{68.0} & \underline{46.5} & 23.0 & \textbf{60.5} & \underline{82.0} & \textbf{37.5} & \underline{93.0} & \underline{50.0} & \underline{78.5} & \textbf{71.0} & \underline{55.7} & \underline{49.0} & \underline{56.9/58.6} \\
V-Random$^\ast$ & 63.0 & 37.5 & 33.5 & 32.0 & 50.5 & 48.0 & 34.0 & 31.0 & 45.0 & 43.5 & 20.0 &57.0 & \textbf{47.0} & \underline{30.5} & 51.5 & 49.5 & \underline{37.0} & 74.0 & 35.0 & 64.5 & 49.2 & 43.4 & 39.3 & 44.2/44.0 \\

\midrule
\rowcolor{graybg}
LLaVA-OV 0.5B & 54.0 & 39.0 & 35.0 & 43.5 & 54.0 & 55.0 & 37.5 & 31.0 & 45.0 & 39.5 & 49.0 & 53.0 & 33.5 & 21.0 & 47.5 & 70.5 & 32.0 & 84.0 & 37.5 & 70.0 & 55.7 & 43.7 & 38.4 & 46.6/45.9 \\
\midrule

\rowcolor{sharpv}

Visual SharpV & \underline{53.5} & \underline{39.5} & \underline{30.0} & \textbf{42.0} & \underline{52.5} & \underline{52.0} & 39.5 & \underline{30.5} & \underline{44.0} & \textbf{37.5} & \textbf{46.5} & \textbf{52.5} & \underline{30.5} & \underline{20.5} & \textbf{48.5} & \underline{69.0} & \textbf{33.5} & \textbf{82.5} & \textbf{39.5} & \textbf{68.0} & \textbf{57.3} & \underline{43.3} & \textbf{39.4} & \textbf{46.3/46.7}\\
V-Random$^\dagger$  & \textbf{55.5} &38.0 & \underline{30.0} & \underline{40.5} &\textbf{55.0} & \textbf{51.5} & \underline{40.0} & \textbf{32.0} & \textbf{45.5} & \textbf{37.5} & \underline{45.0} & 48.5 & \textbf{31.0} & \textbf{22.0} & \underline{48.0} & \textbf{69.5} & \underline{31.0} & \textbf{82.5} & \textbf{39.5} & \underline{67.5} & \underline{55.0} & \textbf{44.3} & \underline{39.2} & \underline{45.5/46.2} \\
V-Random$^\ast$  &\underline{53.5} & \textbf{40.0} & \textbf{31.5} & 24.0  &36.0 & 34.0 & \textbf{40.5}& 27.0 & 37.5 & 37.0 & 25.5 & \textbf{52.5} & 28.0 &20.5 &47.0 &45.5 & 34.0 & 64.0 & 38.0 & 58.0 & 38.8 & 36.4 & 32.3 & 39.1/35.9\\

\bottomrule
\end{tabular}
}
\caption{Ablation Study of Information-aware Threshold in Pre-LLM Stage. $^\dagger$ donates random selection according to information-aware threshold each frame, $^\ast$ indicates random pruning with same threshold of each video sample.}
\label{tab:ablation}
\end{table*}

\begin{figure*}[t]
\centering
\begin{subfigure}[b]{0.33\textwidth}
    \centering
    \includegraphics[width=\linewidth]{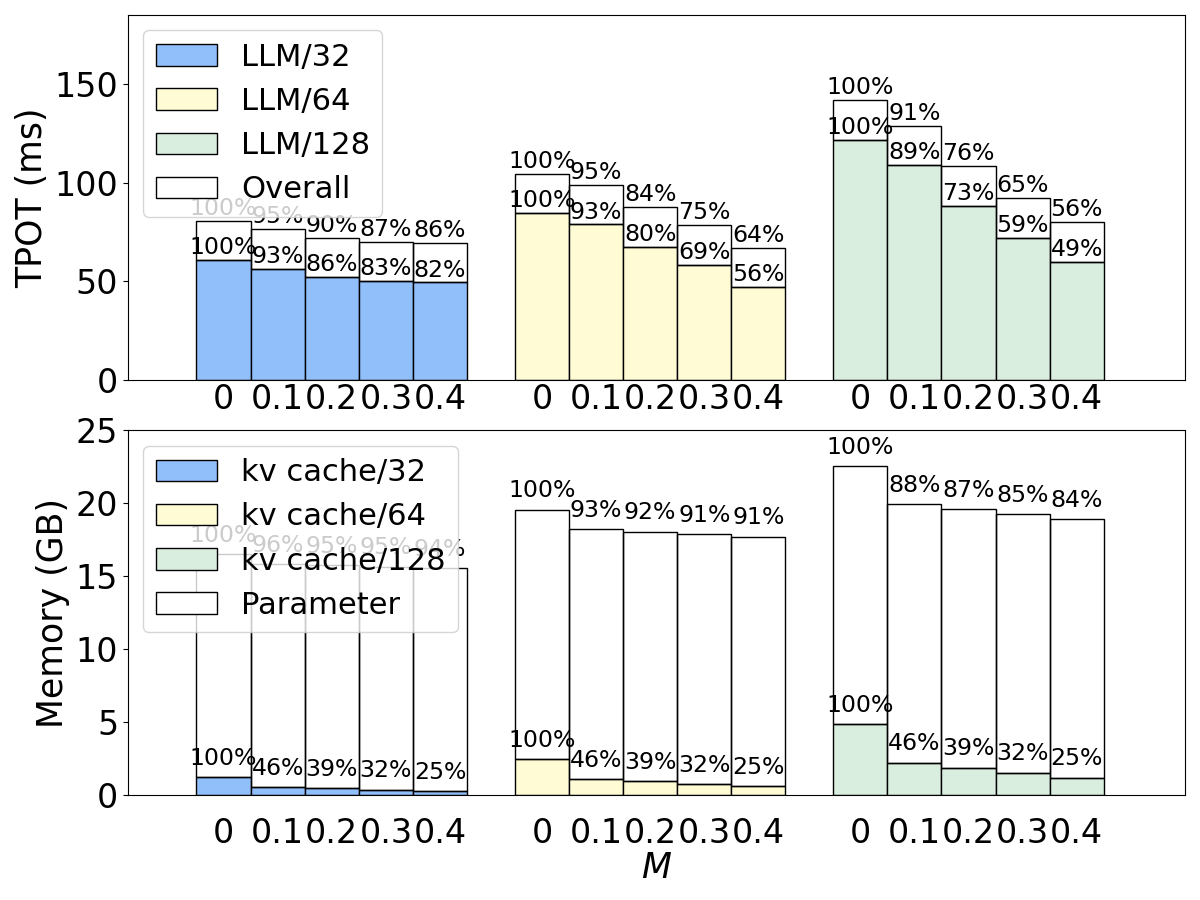}
    \caption{Memory and TPOT Across Frame Sizes}
    \label{subfig:ablation_3}
\end{subfigure}
\hfill
\begin{subfigure}[b]{0.33\textwidth}
    \includegraphics[width=\linewidth]{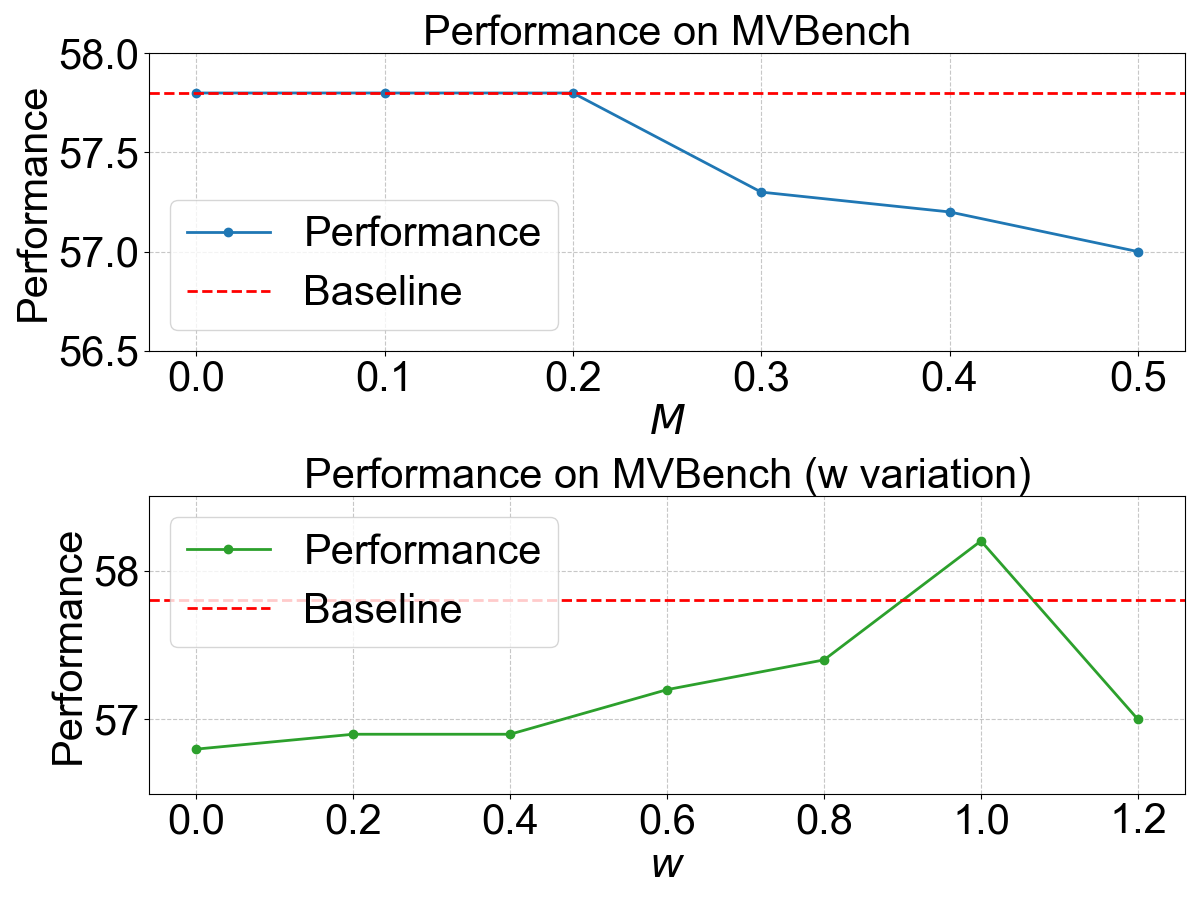}
    \caption{Ablation Study of Parameter $M$ and $w$}
    \label{subfig:ablation_1}
\end{subfigure}
\hfill
\begin{subfigure}[b]{0.33\textwidth}
    \includegraphics[width=\linewidth]{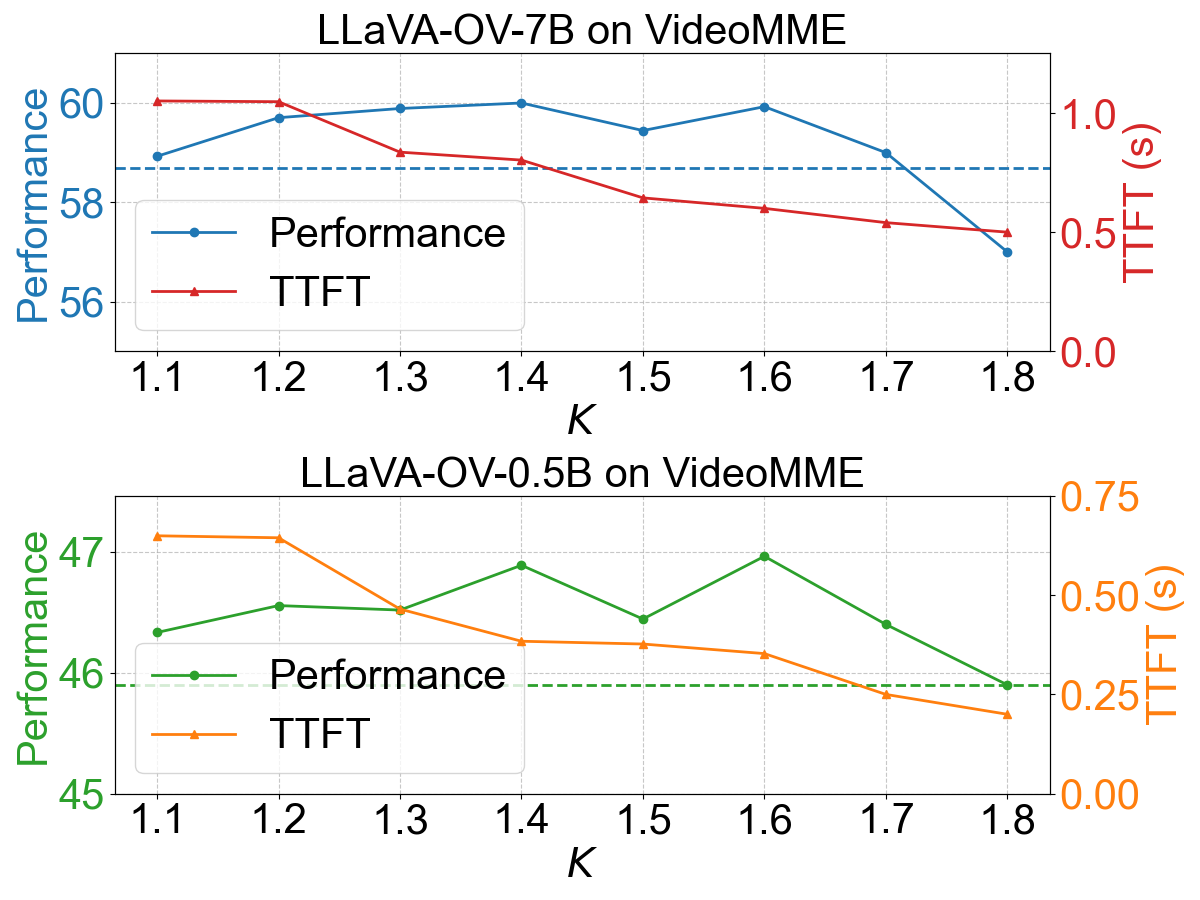}
    \caption{Ablation Study of Parameter K}
    \label{subfig:ablation_2}
\end{subfigure}
\label{fig:ablation123}
\caption{Ablation Study of Parameters: $M,w,K$}
\end{figure*}

\subsection{Performance and Efficiency}

(1) As demonstrated in Table~\ref{tab:main}, SharpV consistently outperforms state-of-the-art methods across various benchmarks while maintaining scalability on both LLaVA-OV-7B and LLaVA-OV-0.5B architectures. Overall, SharpV achieves performance comparable to the dense model with pruning rates between 10\% and 20\% across all four benchmarks, surpassing DyCoke, FastV, and PruMerge while maintaining a lower token budget.

(2) Notably, as shown in Tables~\ref{tab:main} and~\ref{tab:ablation}, SharpV occasionally exceeds the performance of the dense model baseline by 1\% to 2\% on VideoMME, NExTQA, and PLLaVA (VideoMME, MVBench). This improvement can be attributed to the fact that appropriate pruning helps reduce video noise and enables the model to better focus on key visual elements. In contrast, FastV suffers from performance degradation due to its fixed, coarse-grained one-time pruning approach. PruMerge underperforms because it preserves visual encoder attention information that may not align with the LLM's internal attention mechanisms. DyCoke's uniform pruning rate in the first stage results in insufficient preservation of information-rich videos while potentially introducing more noise from information-sparse videos.

(3) Moreover, during the pre-LLM stage, SharpV's adaptive threshold mechanism dynamically retains approximately 32\% of video tokens on average without requiring manual parameter tuning, demonstrating both robustness and generality in achieving optimal performance.

(4) We further compare the multi-dimensional efficiency of different methods in Fig.~\ref{fig:radar} and Table~\ref{tab:main}. While maintaining comparable or superior performance, SharpV achieves significantly lower resource consumption in both GPU memory and FLOPs, attributable to its low computational complexity. Notably, our method demonstrates: 1) reduced TTFT through Visual SharpV's initial pruning stage, and 2) accelerated TPOT enabled by Memory SharpV's subsequent KV cache pruning.

\subsection{Ablation Study}
\subsubsection{Visual SharpV}

In pre-LLM evaluation, we compare our spatial-temporal scoring with two random strategies: 1) V-Random$\dagger$ replaces topK selection with random selection while keeping Visual SharpV's per-frame pruning ratios, comparing score-based versus random selection under identical pruning budgets. 2) V-Random$^\ast$ randomly selects tokens video-wide without frame-wise ratio control, maintaining only the total pruning threshold. Table~\ref{tab:ablation} shows Visual SharpV consistently outperforms V-Random$\dagger$, proving spatio-temporal selection's effectiveness. V-Random$^\ast$ suffers significant degradation versus V-Random$\dagger$, demonstrating the importance of frame-adaptive pruning ratios. Our approach dynamically allocates more tokens to information-rich frames through fine-grained ratio control, better capturing video semantics while reducing redundancy and noise.

\subsubsection{Memory SharpV}
During the intra-LLM stage, we systematically evaluate the KV cache and GPU memory consumption across different frame sizes, as illustrated in Fig.~\ref{subfig:ablation_3}. Our experiments reveal a sharp drop in KV cache usage between 0 and 0.1, where approximately 54\% of visual token layers exhibit similarity scores below 0.1 compared to the original information, while showing stable variations in the $[0.1,0.4]$ range, which align with the analysis presented in Fig.~\ref{fig:simi}. Consequently, TPOT achieves computational acceleration through optimized KV cache reduction.

\subsubsection{Hyper-parameters}
The parameter $w$ controls the degree of spatial information incorporation, accounting for potential distribution differences between intra-frame and inter-frame dissimilarities. As shown in Fig.~\ref{subfig:ablation_1}, optimal performance is achieved at w=1, even surpassing the MVBench baseline. The threshold parameter $m$ determines the degradation boundary. Analysis of similarity degradation in Fig.~\ref{fig:simi} reveals that the decline plateaus when similarity falls within $[0,0.2]$. This observation is empirically validated in Fig.~\ref{subfig:ablation_1}, where performance remains stable for $m <= 0.2$ but deteriorates beyond this threshold. Additionally, threshold $K$ of SharpV(Manual) in Tab.~\ref{tab:main} is decided according to Fig.~\ref{subfig:ablation_2}. A larger $K$ reduces VR and TTFT, with $K=1.6$ offering a balanced efficiency-performance trade-off.

\section{Discussion and Limitations}
SharpV pioneers token-similarity-based dynamic pruning for efficient VLLM inference, yet challenges persist. First, while Visual SharpV’s spatiotemporal compression effectively reduces token redundancy, subtle visual details (e.g., micro-expressions) may still suffer minor information loss. Future work could explore finer-grained token pruning for fine-detail-sensitive scenes. Second, although visual degradation is empirically validated through cosine similarity assisted with information theory and attention shift analysis, a rigorous theoretical framework for deep-layer token transformation mechanisms remains elusive. Further studies could bridge this gap.

\section{Conclusion}
SharpV introduces a novel approach to intra-LLM visual token pruning that operates without reliance on exposed attention scores, achieving linear complexity while maintaining compatibility with Flash Attention. Furthermore, we propose a simple yet effective pre-LLM token selection method, offering a self-calibrated, adaptive thresholding scheme with strong generality and robustness based on information measurement. These components synergize to reduce the final pruning rate around 12\%, while frequently improving accuracy across multiple benchmarks due to reduced visual noise, enhancing both efficiency and reliability in Video-LLM reasoning. 

\section*{Acknowledgements}
This work was supported by the National Natural Science Foundation of China (Grant No.62506318); Guangdong Provincial Department of Education Project (Grant No.2024KQNCX028); CAAI-Ant Group Research Fund; Scientific Research Projects for the Higher-educational Institutions (Grant No.2024312096), Education Bureau of Guangzhou Municipality; Guangzhou-HKUST(GZ) Joint Funding Program (Grant No.2025A03J3957), Education Bureau of Guangzhou Municipality

\bibliography{cite}

\end{document}